\documentclass{article}
\usepackage{ijcai24}

\usepackage{times}
\usepackage{soul}
\usepackage{url}
\usepackage[hidelinks]{hyperref}
\usepackage[utf8]{inputenc}
\usepackage[small]{caption}
\usepackage{graphicx}
\usepackage{amsmath}
\usepackage{amsthm}
\usepackage{booktabs}
\usepackage{algorithm}
\usepackage{algorithmic}
\usepackage[switch]{lineno}

\usepackage{xcolor}
\usepackage{umoline,cuted,bigints}
\usepackage{comment}
\usepackage{amsmath}
\usepackage{amssymb}

\usepackage[acronym]{glossaries}
\makeglossaries

\usepackage{caption}
\usepackage{subcaption}

\usepackage{adjustbox}

\title{Multi-Agent Reinforcement Learning for Energy Networks:\\ Computational Challenges, Progress and Open Problems}

\author{
Sarah Keren$^1$
\and
Chaimaa Essayeh$^2$\and
Stefano V. Albrecht$^{3}$\And
Thomas Morstyn$^4$\\
\affiliations
$^1$Taub Faculty of Computer Science, Technion -- Israel Institute of Technology, Israel\\
$^2$Departement of Engineering, Nottingham Trent University\\
$^3$School of Informatics, University of Edinburgh\\
$^4$Department of Engineering Science, University of Oxford\\
}

\newglossaryentry{prosumer}{
        name=prosumer,
        description={A prosumer is an active end-user with the ability to consume and produce energy.}
}

\newacronym{ai}{AI}{Artificial Intelligence}

\newacronym{ml}{ML}{machine learning}
\newacronym{rl}{RL}{reinforcement learning}
\newacronym{hrl}{HRL}{hierarchical reinforcement learning}
\newacronym{mbrl}{MBRL}{model-based reinforcement learning}
\newacronym{mfrl}{MFRL}{model-free reinforcement learning}
\newacronym{pomdp}{POMDP}{partially observable Markov decision process}
\newacronym{drl}{DRL}{deep reinforcement learning}
\newacronym{dl}{DL}{deep learning}
\newacronym{dqn}{DQN}{deep Q-network}
\newacronym{ddqn}{DDQN}{double deep Q-network}
\newacronym{sac}{SAC}{Soft Actor-Critic}
\newacronym{ac}{AC}{Actor-Critic}
\newacronym{a2c}{A2C}{Advantage Actor-Critic}
\newacronym{ddpg}{DDPG}{Deep Deterministic Policy Gradient}
\newacronym{dpg}{DPG}{Deterministic Policy Gradient}
\newacronym{td3}{TD3}{Twin Delayed Deep Deterministic}
\newacronym{ea}{EA}{Experience Augmentation}
\newacronym{bc}{BC}{Behavioral Cloning}
\newacronym{per}{PER}{Prioritized Experience Replay}
\newacronym{nn}{NN}{neural netrwok}
\newacronym{dnn}{DNN}{Deep Neural Netrwok}
\newacronym{ne}{NE}{Nash Equilibrium}
\newacronym{ppo}{PPO}{Proximal Policy Optimization}
\newacronym{sota}{SoTA}{State of The Art}

\newacronym{mas}{MAS}{multi-agent systems}
\newacronym{marl}{MARL}{multi-agent reinforcement learning}
\newacronym{mdp}{MDP}{Markov decision process}
\newacronym{mapomdp}{MA-POMDP}{Multi-Agent Partially Observable Markov Decision Process}
\newacronym{mamdp}{MA-MDP}{Multi-Agent Markov Decision Process}
\newacronym{eamaac}{EA-MAAC}{Experience Augmented Multi-Agent Actor-Critic}
\newacronym{mg}{MG}{Markov game}
\newacronym{sg}{SG}{stochastic game}
\newacronym{cmg}{CMG}{constrained Markov game}
\newacronym{pomg}{POMG}{Partially Observable Markov Game}
\newacronym{ctde}{CTDE}{Centralized Training with Decentralized Execution}
\newacronym{maddpg}{MADDPG}{Multi-Agent Deep Deterministic Policy Gradient}
\newacronym{matd3}{MATD3}{Multi-Agent Twin Delayed Deep Deterministic}
\newacronym{maac}{MAAC}{Multi-Agent Actor Critic}
\newacronym{masac}{MASAC}{Multi-Agent Soft Actor Critic}
\newacronym{macsac}{MACSAC}{Multi-Agent Constrained Soft Actor-Critic}
\newacronym{oldc}{OLDC}{Online Multi-Agent Reinforcement Learning and Decentralized Control}
\newacronym{mappo}{MAPPO}{Multi-Agent Proximal Policy Optimization}
\newacronym{rdmrl}{RDMRL}{Recurrent Deep Multi-Agent Reinforcement Learning}
\newacronym{dial}{DIAL}{Differentiable Inter-Agent Learning}

\newacronym{dedp}{DEDP}{Dynamic Economic Dispatch Problem}
\newacronym{uc}{UC}{Unit Commitment}
\newacronym{dr}{DR}{Demand Response}
\newacronym{dsm}{DSM}{Demand-Side Management}
\newacronym{sd}{SD}{supply demand}
\newacronym{dem}{DEM}{distributed energy management}
\newacronym{pv}{PV}{photovoltaic}
\newacronym{res}{RES}{renewable energy sources}
\newacronym{der}{DER}{distributed energy resources}
\newacronym{ess}{ESS}{energy storage system}
\newacronym{lv}{LV}{Low Voltage}
\newacronym{vvc}{VVC}{Volt/Var Control}
\newacronym{avc}{AVC}{Active Voltage Control}
\newacronym{mv}{MV}{Medium Voltage}
\newacronym{ev}{EV}{electrical vehicles}
\newacronym{vpp}{VPP}{Virtual Power Plants}
\newacronym{mpc}{MPC}{Model Predictive Control}
\newacronym{rbc}{RBC}{Rule-Based Controller}
\newacronym{soc}{SoC}{State-of-Charge}
\newacronym{zne}{ZNE}{Zero-Net Energy}
\newacronym{dt}{DT}{Digital Twin}
\newacronym{kpi}{KPI}{Key Performance Index}
\newacronym{pf}{PF}{Power Flow}
\newacronym{spgp}{SPGP}{Sparse Pseudo-Gaussian Process}
\newacronym{hem}{HEM}{Home Energy Management}
\newacronym{rtp}{RTP}{Real Time Pricing}
\newacronym{tso}{TSO}{Transmission System Operator}
\newacronym{gm}{GM}{Grid Manager}
\newacronym{ga}{GA}{Grid Agent}
\newacronym{td}{TD}{Temporal Difference}

\newacronym{gem}{GEM}{Grid Edge Management}
\newacronym{psoc}{PSOC}{Power System Operation and Control}
\newacronym{em}{EM}{Electricity Market}
\newacronym{lb}{LB}{Load Balancing}
\newacronym{fc}{FC}{Frequency Control}
\newacronym{ed}{ED}{Economic Dispatch}

\newcommand{\states}{
\mathcal{S}}

\newcommand{\actions}{
\mathcal{A}}

\newcommand{\transFunc}{
\mathcal{T}}

\newcommand{\rewardFunc}{
\mathcal{R}}

\newcommand{\jointRewardFunc}{
\rewardFunc}

\newcommand{\discount}{
\gamma}

\newcommand{\actionn}{
a}

\newcommand{\state}{
s}

\newcommand{\policy}{
\pi}

\newcommand{\jointPolicy
}{
\policy}

\newcommand{\jointActions}{\mathcal{A}}

\newcommand{\jointAction}{a}

\newcommand{\observations}{O}
\newcommand{\observation}{o}
\newcommand{\sensorFunc}{\mathcal{O}}

\newcommand{\belief}{b}

\newcommand{\beliefs}{\mathcal{B}}
\newcommand{\Value}{V}

\newcommand{\EnergyNetwork}{energy network}

\newcommand{\RES}{RES}
\newcommand{\DER}{DER}

\usepackage{comment}

\newcount\Comments
\usepackage{xcolor}
\definecolor{darkgreen}{rgb}{0,0.7,0}
\newcommand{\kibitz}[2]{\ifnum\Comments=1{\color{#1}{#2}}\fi}
\newcommand{\sk}[1]{\kibitz{teal}{[sk: #1]}}
\newcommand{\tm}[1]{\kibitz{blue}{[tm: #1]}}
\newcommand{\sa}[1]{\kibitz{purple}{[sa: #1]}}

\newcommand{\todo}[1]{\kibitz{darkgreen}{[todo: #1]}}
\newcommand{\remove}[1]
{\kibitz{yellow}{[remove: #1]}}

\begin{document}

\maketitle

\begin{abstract} The rapidly changing architecture and functionality of electrical networks and the increasing penetration of renewable and distributed energy resources 
have resulted in various technological and managerial challenges. These have rendered traditional centralized energy-market paradigms insufficient due to their inability to support the dynamic and evolving nature of the network.
This survey explores how {\em multi-agent reinforcement learning} (\acrshort{marl}) can support the decentralization and decarbonization of energy networks and mitigate the associated challenges. This is achieved by specifying key computational challenges in managing energy networks, reviewing recent research progress on addressing them, and highlighting open challenges that may be addressed using \acrshort{marl}.

\end{abstract}
\section{Introduction}\label{sec:intro}

Recent technological advancements have given rise to {\em smart grids}, electricity networks in which novel power generation, storage, and information technologies are used to monitor and manage the production, consumption, and transmission of electricity within an electrical network \cite{nair2018multi,CHARBONNIER2022119188,ZHU2023120212,CAPPER2022112403}. Together with their potential to increase efficiency, reduce resource waste and costs, and maximize the transparency and reliability of energy supply, these technological advancements have resulted in the electrical grid's rapidly changing architecture and functionality and have given rise to a wide range of technical and managerial challenges.

Among the key challenges is the increasing penetration of \emph{\acrfull{res}} such as solar and wind energies, which increased in volume in the last several years with the acceleration of \acrfull{pv} solar panels installation in private properties, and the rapidly increasing market of \acrfull{ev},%
\tm{photovolatic is usually one word I think}
which changed the demand and distribution patterns within the network and have introduced
many sources of uncertainty since their power output is uncertain (hard to predict), intermittent (exhibits large
fluctuations), and largely uncontrollable (not dispatchable on command) \cite{kamboj2011deploying,hu2015multi}.

The decentralization of supply and demand raises the need to find novel ways to manage the electrical grid at two highly correlated levels of abstraction. %
The first focuses on maintaining the grid's electrical quality and stability. This requires simultaneously considering a large number of consumers and producers and
maintaining the supply-demand balance at fast time-scales and managing frequency, voltage, and power flow limits throughout the system. %
The second deals with the management and regulation of {\em electrical markets} in which various producers and consumers trade electricity.%

From a computational perspective, the grid's transition to a decentralized architecture has a tremendous effect on its management and decision-making and renders the traditional centralized methods insufficient \cite{fahnrich2015multi,roche2010multi}. %
Even if all components of the system could be controlled by one entity, centralized decision-making is highly inefficient since it requires a large spread of metering devices that continuously communicate their measurements to the centralized controller and requires high-volume data flow in order to support optimal decision-making. Moreover, due to the scale and complexity of the network, %
traditional methods for examining the grid's stability %
become intractable.

We review methods for replacing centralized control with distributed and hierarchical models in which autonomous and semi-autonomous agents represent system components. %
We focus on \emph{\acrfull{marl}}  
\cite{marlbook} in which multiple agents %
learn from their interaction with the environment and with one another while aiming to maximize their expected utilities. 
As opposed to many common virtual MARL settings, the challenge here is in optimizing the behaviors of grid agents while considering the constraints and optimization considerations imposed by the physical electrical network.

\begin{figure*}
     \centering

     \includegraphics[width=0.85\textwidth]{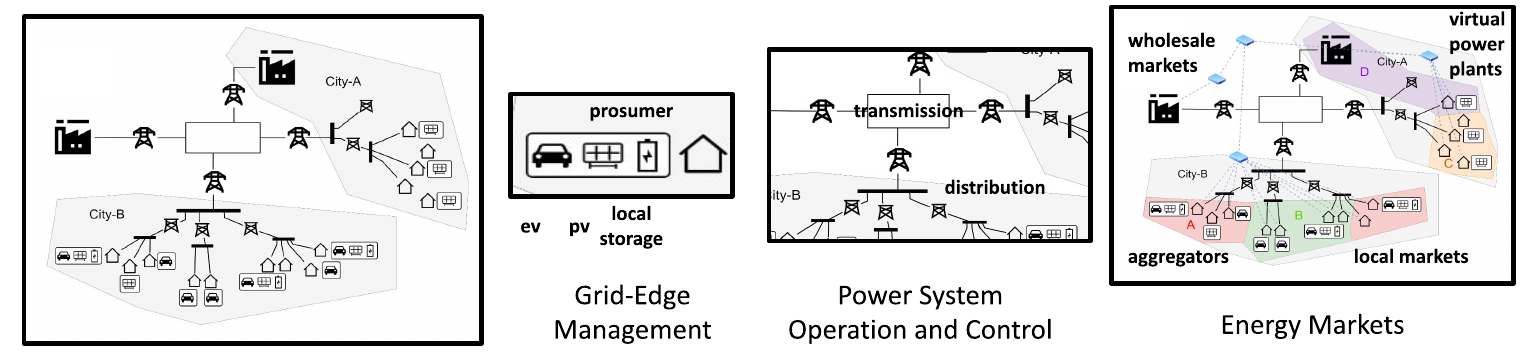}
        \caption{Energy network challenges.
        From left to right:
        (a) an example energy network
        (b) optimizing the policy of a single prosumer: Section \ref{sec:GEM}
        (c) optimizing the operation of the electrical grid: Section \ref{sec:psoc}
     (d) creating new energy markets structures: Section \ref{sec:EM}.}
        \label{fig:categories}
\end{figure*}

While several surveys review the use of \acrshort{marl} for energy networks (e.g.,  \cite{mahela2020comprehensive,CHARBONNIER2022119188,SCHWIDTAL2023113273,ZHU2023120212,capper2022peer}), %
this is, to the best of our knowledge, the first that is aimed at AI researchers and at highlighting the challenges they can help address. For this, we specify some of the key computational challenges of modern energy network management (Section \ref{sec:challenges}), provide an overview of some of the current \acrshort{marl} methods for solving them (Section \ref{sec:solutons}) and specify open challenges yet to be addressed toward the successful integration of \acrshort{marl} for energy networks (Section \ref{sec:gaps}).

\vspace{-.15cm}

\section{Key Computational Challenges in Energy Network Management}\label{sec:challenges}
\remove{electrical network is a complex system comprised of different entities with different functionalities. Generation involves the production of electricity or other forms of energy from various sources. Transmission involves the transportation of electricity or other energy resources from the point of generation to a distribution network, which involves the delivery of energy to end-users, such as homes, businesses, and industrial facilities that %
purchase and consume energy for various purposes. 
In addition, the network includes entities that manage and regulate the energy market and its operation. }

Traditionally, each entity in the electrical network had a single functionality and the network was centrally managed and controlled \cite{ZHU2023120212}. 
The recent introduction of new production and storage technologies and the emergence of new and uncertain consumption patterns have led to the grid's decentralization and have given rise to many computational challenges in the %
optimization of its operation at different scales.
We aim to highlight some of the key challenges that are especially suitable to be addressed using MARL.

We classify the presented problems using three main categories (see Figure \ref{fig:categories}). Section \ref{sec:GEM} considers the point of view of a single electrical component, that may either represent a single or set of devices (e.g., household)  
that interact with the 
network as one entity to optimize consumption, storage, and production. %
Section \ref{sec:psoc} considers a network manager and its need to control the electrical grid to maintain its stability. %
Section \ref{sec:EM} broadly considers the management of the different energy markets in which energy and flexibility services are traded among different utility-maximizing entities. 

We note that categorizing the challenges of energy network management is not straightforward. Many issues are interconnected and their solution is often combined. For example, ensuring that the physical system is stable requires predicting the energy consumption of various users, adjusting production, and managing electricity prices. However, once prices are set, they affect consumption, which may cause instability and raise the need to readjust the system, and so on. With modern \acrshort{res}s and local storage units, maintaining market and network stability becomes even more challenging.
\sa{it might be useful to have an image that summarises/shows the components from the subsections (GEM, PSOC etc) and their relation; e.g. block diagram or something else?}

\subsection{\acrfull{gem}}
{\bf}\label{sec:GEM}
\acrfull{gem} considers energy usage of grid-edge entities. Within this realm, recent technologies have created a major shift in the management of energy consumption within a household, referred to in the literature as \acrfull{hem}. 
As opposed to traditional homes which were passive consumers that only included energy-consuming appliances such as lighting appliances, washing machines, and air conditioners,
modern homes have now become ‘prosumers’ – consumers who proactively manage their consumption, production and storage of energy.
This includes managing heating and cooling systems, \acrshort{ev}s, and local storage units. Thus, methods for managing a modern home or other grid-edge entities need to consider a combination of sensors, communication devices, and control algorithms that can monitor and control energy usage. 
The primary objective is to optimize energy consumption by minimizing cost (and sometimes carbon emission) while respecting usage requirements such as comfort.

An important characteristic of these settings is the electrical network and the details of its operation are typically abstracted and simplified. Instead, it is considered a part of the environment with which the grid-edge entities interact. 
Thus, for example, if a prosumer decides to turn on an appliance and increase its consumption of electricity it is typically assumed that this requirement can be satisfied and the network's need to adjust to this new requirement is abstracted.

\subsection{\acrfull{psoc}}\label{sec:psoc}
Here we consider the set of activities and strategies employed to ensure the reliable, stable, and efficient operation of an electrical power network. 
Power in the network is divided into {\em real power}, or {\em active} power, which represents the actual energy transferred and consumed in an electrical circuit to perform `useful work' (e.g., mechanical motion, heating, or light) and is the power that is bought and sold in electricity markets %
while {\em reactive power}, %
does not perform useful work but plays a crucial role in maintaining voltage levels in the electrical grid and in supporting its stability and efficiency.

To maintain system stability and reliability, efficient management of both real and reactive power is essential.
Accordingly, \acrfull{psoc} includes various challenges that are related to both aspects including \acrfull{lb}, which focuses on the distribution of electrical loads across different generators and transmission lines such that no individual component is overloaded, \acrfull{pf}, which deals with the analysis of the steady-state behavior of a power system and the distribution of real power within the network, \acrfull{vvc} which is related to the regulation of reactive power and is aimed at optimizing power quality, reducing energy losses, and ensuring grid stability, \acrfull{fc} which involves maintaining the power system frequency at the standard value by adjusting the generation and load, and more. 

A key barrier to the integration of \acrshort{der}s to the electrical grid is the challenges they pose on the operation and control of the system.
This is due to their intermittent and variable generation patterns, which can lead to uncertainty in power supply and unexpected effects on voltage and frequency levels\remove{This makes it challenging to maintain voltage and frequency within acceptable limits,} %
and their requirement for bidirectional power flow from and to the distribution network. Moreover, although 
\acrshort{der}s can locally supply energy during a grid outage, their integration impacts the traditional protective devices and stability mechanisms designed for traditional power generation and poses safety risks that may hinder restoration efforts. 

The effective integration of \acrshort{der} requires novel control strategies to manage the variability and intermittency and ensure grid stability, advanced methods for forecasting generation and demand, and new methods for the real-time monitoring and control of the network. \sa{could add some cites to this and prior section?}

 \subsection{\acrfull{em}}\label{sec:EM}
 
In an \acrfull{em}, electricity is a commodity traded between participants that can generate, store, and consume electricity.  
These can be generally characterized as {\em regulated markets}, in which a single authority (e.g., a government)  controls the different aspects of electricity generation, distribution, and pricing, and {\em deregulated markets}, or {\em competitive markets}, in which multiple electricity providers compete for consumers and market forces often determine prices.

The recent shift towards decentralized energy systems,
in which communities of grid entities can satisfy more of their own energy needs from renewable energy generated from local sources and in which information technologies facilitate information flow \cite{cremers2022efficient}, has raised the need for novel structures of deregulated markets.
Such markets include two levels of abstraction: wholesale and local.

In a {\em wholesale 
 market} electricity is traded before being delivered to consumers and includes {\em generators}, which produce energy, and {\em suppliers} which are tasked with meeting consumer demands at any given time.
 Due to the uncertainty in supply and demand, several markets are typically maintained for different time horizons, i.e., Real-Time, Day-Ahead, etc \cite{bose2019some,ZHU2023120212}.
 
 \remove{
Typically, a deregulated market is managed by an {\em Independent System Operator (ISO)} that manages the transmission grid and ensures fair market access for all market participants.
}

The basic setting is one in which the consumption profile of the various grid entities as well as their production and storage capabilities are fixed, and the objective is to maintain grid stability by equalizing supply and demand at all times while minimizing operational costs. %
This problem is typically studied as \acrfull{uc} that considers which generating units in a power system should operate over a specific time horizon (typically, days) %
while considering factors such as generation limits, startup, and shutdown costs, minimum up and down times, and other operational constraints. 
Strongly related to \acrshort{uc} is the problem of \acrfull{ed} which considers shorter horizons (typically real-time or several minutes) and the objective is to allocate the power output %
to meet the system demand at the lowest possible cost.

In more advanced settings, the reward-maximizing and strategic nature of grid agents is accounted for and it is possible to influence their behavior using monetary incentives and penalties. One relevant problem is 
\acrfull{dr}, or demand-side management, which focuses on managing electricity demand by incentivizing customers to adjust their consumption in response to changing grid conditions. This involves designing incentives to encourage specific customer behaviors, as well as developing approaches for predicting demand patterns and coordinating customer responses.
A typical goal is to promote stability by encouraging customers to reduce energy use during high-demand periods, such as hot summer days when air conditioning usage is high. 

In contrast to wholesale markets which typically consider large entities, %
 {\em local energy markets} consider smaller entities and the interaction between suppliers and end-users (residential and business consumers). Electricity traded in this market includes both that procured from generators in the wholesale electricity market and electricity generated by local generators 
\cite{CAPPER2022112403,cremers2022efficient}.
 Various models have been offered for organizing \EnergyNetwork s as {\em transactive
energy systems}, in which economic and control mechanisms dictate the dynamics of the system using utility as a key operational parameter \cite{cremers2022efficient,robu2012cooperative}. In general, these typically include a set of DERs, an interconnecting local network, an upstream energy market, and a digital coordination platform for sensing, communications, and control \cite{CHARBONNIER2022119188}.

An important and challenging characteristic of modern energy networks is that novel technologies enable interactions that can be used not only to support the local sharing of physical resources but also to support interactions of spatially separated entities (e.g., households in different cities) that are only connected by virtual agreements \cite{robu2012cooperative}.
Thus, local communities of anywhere between several units to several hundred houses (e.g. a village or a city neighborhood) can be formed to share energy assets, such as wind turbines or storage units, as well as the energy bill for the aggregate residual demand, i.e., the part of the demand not covered by the local generation and storage assets. At the same time, these technologies support the formation of {\em virtual power plants} by which dispersed generation and consumption units can form an aggregate that may even be large enough for participating in wholesale markets as 
large-scale producers and consumers. Here, finding optimal groupings within the large network and forming sustainable agreements becomes a computational challenge.

Another important challenge arises from the ability to exploit {\em power flexibility} as a commodity that can be traded \cite{10.1145/3360322.3360998}. %
Power flexibility can be regarded as the capability to reduce, increase, or shift electricity consumption or generation in response to an economic signal. This capability is gaining importance as a way to offset energy imbalances and help manage network constraints. \remove{State-of-the-art approaches for optimal coordination of grid-edge flexibility have focused on convex relaxations and decomposition techniques, but these fail to support systems of realistic scale. Another set of approaches has focused on aggregated coordination based on {\em flexibility envelopes} \cite{nosair2015flexibility}, which enables tractable integration of grid-edge flexibility into transmission system optimization but does not capture grid-edge flexibility at the distribution network level.}

\remove{
One prominent model is {\em peer-to-peer} (P2P) energy trading, where grid entities can trade electricity with their neighbours directly, based on their private assets (solar PV panels, batteries, etc). %
The second model considers the formation
of {\em energy communities}, where grid entities group together to commonly own and share
energy assets and interact with the utility grid and the grid manager as a unified entity. 
Many such community models include a {\em community aggregator}, which controls
the energy assets and distributes the generated/discharged power to
members of the community \cite{morstyn2018using}. 
The aggregator is also in charge
of receiving energy from the utility grid whenever there is a deficit.
The new technologies have enriched the set of possible interactions within the retail market and offer novel ways of collaboration.}

\vspace{-.2cm}
\section{MARL for Energy Network Management}\label{sec:solutons}
Many recent frameworks address the problems described in Section \ref{sec:challenges} using \acrshort{marl}, which is considered here as a form of (\acrshort{rl}), wherein multiple agents learn to optimize their accumulated rewards while interacting with their partially known environment and with other agents. Due to space constraints, we exclude a detailed formal account of MARL and its solution methods and refer the reader to \cite{marlbook}.

 The most commonly utilized \acrshort{marl} model is the Stochastic Game (\acrshort{sg})  (also known as Markov Game or Multi-agent MDP) \cite{shapley1953stochastic} defined as a tuple
 $\langle \states, \jointActions = \lbrace\actions_i\rbrace_{i=1}^{n}, 
\transFunc,\jointRewardFunc = \lbrace\mathbf{\rewardFunc}_i\rbrace_{i=1}^{n}, \discount\rangle$, 
where $\states$ is the state space, 
$\jointActions$
is the 
{\em joint action space} with $\actions_i$ as the $i^{th}$ agent action space s.t.   $\jointAction\triangleq\left(\actionn_1, \actionn_2, \dots, \actionn_n\right)$ for $\jointAction \in \jointActions$, $\transFunc:\states\times\jointActions\times\states\rightarrow [0, 1]$ is the transition probability function  $\transFunc\left(\state', \jointAction, \state\right)$  such that $\forall \state \in \states, \forall \jointAction \in \actions: \sum_{\state' \in \states} \transFunc(\state, \jointAction, \state') =1$,   %
$\jointRewardFunc$ is the 
{\em joint reward function} with $\rewardFunc_i: \states \times \jointActions \times \states \rightarrow \mathbb{R}$ as the $i^{th}$ agent reward function,
and $\gamma$ is the discount factor. 
A solution to an \acrshort{sg}~is a joint policy $\jointPolicy\triangleq (\policy_1, \dots, \policy_n)$ which associates each agent with a policy
$\policy_i:\states\times \actions_i\rightarrow [0, 1]$ that specifies the probability of agent $i$ taking an action at a given state.
\remove{
Each joint policy $\jointPolicy$ and agent $i$ is associated with a value function $\Value^{i}_{\jointPolicy}(\state)$ that represents the expected accumulated re agent $i$ will receive when the joint policy $\policy$ is followed from state $\state$ onward, 
and a 
{\em $Q$-function} $Q^{i}_{\jointPolicy}(\state,\jointAction)$ that represents the expected accumulated reward of agent $i$ after joint action $\jointAction$ is performed at state $\state$ and $\jointPolicy$ is jointly followed thereafter \cite{sutton2018reinforcement}
} In partially observable settings, the definition also includes for each agent $i$ a finite set of observations ${\observations}_i$ it may observe %
and a sensor function  $\sensorFunc_i: \jointActions \times  \states\times\observations_i \rightarrow [0, 1]$ which determines the probability $\sensorFunc_i(\jointAction, \state, \observation)$ it perceives observation $\observation\in\observations_i$  at state $\state \in \states$ state after joint action $\jointAction \in \jointActions$ is performed (alternatively defined as the conditional probability $\sensorFunc_i(\jointAction| \state, \observation)$)  \cite{hansen2004dynamic}. 
Here, the (joint) policy is defined over (joint) {\em beliefs} $\beliefs$ that represent the probability of each state. 
 
\remove{. Each belief $\belief\in \beliefs$ is defined as $\belief \triangleq (\belief_1, \dots, \belief_n)$} %
\remove {specifying the probability  $\belief_i(\state)$ agent $i$ associates to state $\state$ being the true state. }\sk{add something about histories?}

This general definition captures a variety of interactions and relationships that can exist between agents in collaborative, competitive, and mixed-incentive MARL settings.
The complex interactions among agents may give rise to behaviors that are difficult to anticipate by simply examining each agent in isolation.  
Thus, despite the potential to solve complex problems across various domains, \acrshort{marl} faces various significant challenges  \cite{marlbook}. These stem
from aspects such as scale, conflicting goals of self-interested agents, different partial views of the environment, the fact that agents are concurrently learning to optimize their policies while causing the environment to be non-stationary from the perspective of each agent, and the {\em credit assignment} challenge which involves determining which action and agent 
contributed to received rewards. 
All these are relevant to MARL settings in general but are particularly relevant to energy networks with the added need to account for the dynamics of the physical environment and the effect decisions may have on the functioning of the electricity network.

\remove{
All the mentioned considerations are relevant to MARL settings in general but are particularly relevant to our settings or interests, in which there is the added complexity associated with the need to account for the dynamics of the physical environment and the effect decisions may have on the functioning of the electricity network.
Thus, relying solely on model-free methods can be ineffective due to data scarcity and the uncertain and dynamic nature of the environment. At the same time, the complexity of the settings makes it hard to fully rely on models of the dynamics of the electrical entities in the system that may not be available or accurate. A promising direction is therefore the use of combined methods that exploit the benefits of both model-based and model-free approaches.
}

Hereon, we describe traditional ways for solving each of the challenges mentioned in Section \ref{sec:challenges} and how they can be modeled and solved using \acrshort{marl}.  %
Notably, our account of related work is not comprehensive and is instead aimed at providing examples of research on MARL for energy systems and at setting the path to developing novel MARL methods for addressing the many unresolved challenges.%

\vspace{-.2cm}
\subsection{\acrfull{gem}}
\remove{This relatively new concept gained importance with the emergence of DERs and the deployment of novel AI technologies and storage units in domestic settings.} %
Although \acrshort{gem} and \acrshort{hem} settings typically consider relatively small-scale settings, traditional methods that mostly rely on linear programming  \cite{ucctuug2012linear,amini2015smart} have become insufficient. This is because they are unable to account for the various forms of uncertainty related to the new technologies, such as the inability to accurately predict the generation patterns of PVs, and the need to adapt to dynamic pricing in contrast to traditional Time of Use (ToU) prices that are known in advance.

To achieve optimal performance and fully benefit from the domestic asset portfolio,
\acrshort{gem} methods are required to consider long-horizons
and to adapt and change their policy in real time. For example, the optimality of the decision of whether to store electricity in a local storage unit or sell it to the grid depends on a prediction of electricity prices and consumption volumes and on the ability to quickly adapt to unexpected events, such as major power failures.

\paragraph{\acrshort{gem} as \acrshort{marl}:} 
A common formulation for \acrshort{gem} and \acrshort{hem} considers a group of electrical entities within a grid-edge unit or household as strategic utility maximizing agents \cite{fangMultiagentReinforcementLearning2020,charbonnierScalableMultiagentReinforcement2022,jendoubi2023multi}. The action space $\actions_i$ of each appliance represents its ability to schedule and control the volume of power allocation for their appliance. The state space $\states$ captures the set of power allocation schedules (e.g., electric vehicle charging, then operation the washing machine), the grid parameters (e.g., electricity price, carbon footprint, etc.), and environment and weather parameters (e.g., temperature, solar radiance, humidity, etc.). The reward function $\rewardFunc_{i}$ of the $i^{th}$ appliance is typically given by:
\begin{align}\label{formula:reward}
    \rewardFunc_{i}(s) &= (1-\lambda)F^{S}_{i}(s) + \lambda F_{i}^C(s),\quad \lambda\in [0, 1],    
\end{align}
where $F^{S}_{i}(s)$ refers to the user-defined satisfaction level at state $s$ (e.g., comfort), while $F_{i}^C(s)$ is the consumption cost at state $s$, which depend on the functionalities of the different appliances. The objective is to identify an optimal control policy that maximizes the collective long-term return of all device agents within the shared environment.

\paragraph{Example Solution Approaches:}
The \acrshort{gem} is a relatively small-scale setting that typically involves a single optimization objective of the grid-edge entity. For example, when used to optimize a household it includes a relatively small number of appliances with the single objective of minimizing the total utility expenses. %
This means that in principle such a setting can be solved by adopting a centralized fully observable model by which a single RL agent optimizes its policy defined over the joint action space of all appliances. 
However, since each electrical appliance may be associated with several sensors and there may be a need to consider many diverse parameters and optimization considerations (e.g., cost and carbon emission), it may become beneficial from a computational perspective to find hierarchical and decentralized control methods for such settings.

As an example, in
\cite{xuMultiAgentReinforcementLearningBased2020} a \acrshort{sg} framework for \acrshort{hem} is offered which involves four types of agents that are responsible for controlling four distinct categories of appliances: \emph{non-shiftable}, i.e., appliances that offer no flexibility in their operation schedules and for which demand must be met at all times (e.g., medical devices, light bulbs), \emph{power-shiftable}, i.e., appliances that can adjust their power consumption (e.g., air conditioner), \emph{time-shiftable}, i.e., appliances that can shift their time of operation but cannot be interrupted during their functioning (e.g., washing machine and dishwasher), and appliances that can be considered both time-shiftable and power-shiftable, including \acrshort{ev}s for which both charging schedules and power levels can be controlled. With the expectation of \emph{non-shiftable} appliances that do not have agency, each appliance is associated with an agent that determines the hour-ahead energy
consumption of its appliance and an individual reward function that considers costs, dissatisfaction penalties (e.g, waiting time), etc.
The suggested \acrshort{marl} method involves a shared neural-network-based learner (Feedforward NN) that predicts future electricity prices and solar generation and a decentralized Q-learning approach for optimizing the operation of each appliance.

As another example, the setting in \cite{leeDemandSideSchedulingBased2020} 
 involves multiple households equipped with a smart
meter that can schedule appliances online, based on user tasks (e.g., laundry cycle) that appear according to some modeled distribution. The objective of each household is to optimize the individual accumulated reward which is a linear combination of the cost and the delay in performing the incoming tasks. 
The proposed \acrshort{sg}~formulation 
adopts a practical assumption that each household can only observe its own internal state along with the published electricity price at the previous time step. This supports scalability and privacy-preservation in a realistic setting. 
The solution approach is based on the commonly used  {\em \acrfull{ctde}} \cite{lowe2017multi} approach by which agents jointly learn a shared {\em critic} that optimizes the Q-value that is associated with each state and action but adopt an individual {\em actor} that controls the executed policy.

\subsection{\acrfull{psoc}}

A key characteristic of the various challenges of \acrfull{psoc} is that they require accounting for various constraints imposed by the physical electrical networks. 
Accordingly, most traditional solution approaches rely on methods that can incorporate multiple constraints within the analysis and optimization process including numerical algorithms, linear and non-linear programming, and control theory. For example, the widely used methods for solving the \acrshort{pf} equations are the Newton-Raphson and Gauss-Seidel methods which iteratively calculate bus voltages and phase angles to achieve a balanced system. Similarly, dynamic programming and Model Predictive Controllers (MPCs) are used for settings that require long-horizon decision-making such as optimal generation scheduling and energy storage management. %

To account for the recent technological advancements and shifts in the network structure, new methods have been offered that apply different relaxations and heuristics and offer robust control techniques that can handle uncertainties and variations in power system parameters. 
For example, DC Power Flow (Linearized Power Flow) achieves computational efficiency by neglecting reactive power and assuming bus voltage magnitudes are bounded. %

Another leading approach involves the decomposition of the power system into subsystems, allowing for parallel processing and computation. While these improve scalability and efficiency, they require careful handling of boundary conditions between subsystems.
To demonstrate, a key aspect of \acrshort{vvc} involves the control of capacitor banks and tap changers to keep the local voltage within an acceptable range. Such local decisions may have a sub-optimal effect on the rest of the network as adjusting the VAR value at one point of the electrical network can have negative impacts on the value at a neighboring point. 
Adding to this complexity in modern smart grids are the new devices and \acrshort{res}s connected to local distribution networks. While these can control their reactive power and participate in \acrshort{vvc} they introduce many challenges due to their high generation and consumption uncertainty. This is added to the need to deal with scalability issues of controlling voltage in large networks, and the need to account for the locational interdependencies and frequent fluctuations that may occur unexpectedly, which is highlighted in modern electrical networks. All these have rendered the mostly rule-based traditional methods (e.g., \cite{uluski2011volt,bollen2015volt}) insufficient.

A key strength of \acrshort{marl} models and methods for \acrshort{psoc} is their ability to %
support decentralized control strategies and effective distributed computations across different components or agents in the power system \cite{gaoConsensusMultiAgentReinforcement2021}. Thus, even though \acrshort{psoc} does not typically involve strategic agents, such as in the energy markets, %
it is sensible to consider \acrshort{marl} as a way to achieve computational efficiently for problems that are challenging for conventional methods.

One such challenge is \emph{active voltage control} (AVC) which refers to exploiting devices connected to an electrical network to regulate and maintain its voltage levels within desired limits. 
Voltage control problems have been studied for many years, but have recently come under the spotlight due to the increasing penetration of \acrshort{res}, and 
 \acrshort{pv}s in particular, that may cause voltage fluctuations. Since these new devices appear in vast numbers and are distributed across wide geographic areas, it is possible to exploit their control flexibility, together with the flexibility of other controllable devices, such as Static Var Compensators (SVCs) and On-Load Tap Changers (OLTCs) to regulate voltage throughout the network. 
Since AVC requires accounting for both local and global parameters and for the propagated influence decisions of nodes have on their connected nodes, and since it involves relatively less severe consequences in the case of failures and constraint violations, \acrshort{marl} is a suitable and promising solution approach \cite{wang2021multi}.

\vspace{-.1cm}

\subsubsection{\acrshort{psoc} as a \acrshort{sg}}

A typical model for various forms of \acrshort{psoc} analysis is a graph $\mathcal{G}=(\mathcal{V}, \mathcal{E})$, where the set of nodes $\mathcal{V}$ represent the electrical components and the edges represent the relationships between them. For example, for \acrshort{vvc} the nodes represent buses, i.e., components that facilitate the transfer of electrical energy between different components of the network, and edges $\mathcal{E}\triangleq\mathcal{V}\times \mathcal{V}$ represent the connectivity between buses in a distribution network. The graph is also typically associated with an {\em admittance matrix} where  $\mathbf{Y}_{N\times N}$ represents the electrical current that can flow between nodes. %

The network graph is used to represent the relationships between agents in a \acrshort{sg}. As an example, in \cite{wang2021multi} a Decentralized Partially-Observable MDP (Dec-POMDP) \cite{bernstein2002complexity,oliehoek2016concise} is offered for modeling and solving the AVC Problem for a network of \acrshort{pv}s. A Dec-POMDP is a special case of a partially observable \acrshort{sg} in which agents are collaborating to maximize a shared reward. In this representation, adopted from \cite{gan2013optimal}, the medium and low voltage distribution networks \sk{explain?} are modeled as a \sk{tree} graph where the nodes $\mathcal{V}$ are the \acrshort{pv}s and the edges $\mathcal{E}$ are characterized by the active and reactive power injection formulas associated with the connection (i.e., transmission line) between them. 

The \acrshort{sg} formulation associates each node with an agent that controls the generation and absorption of reactive and active power, as dictated by its capacity and safety constraints. The state space represents the nodal features that are needed to compute the stable state of the network and include the set of active and reactive power of loads (consumed by the node), the active and reactive power generated by the \acrshort{pv}s, and the voltage magnitude and phase. Observations include the current nodal state and possibly the state of the neighbors, assuming that each agent can communicate and share its local information with its neighbors defined by the network /\ communication graph $G$.

To reduce computational complexity, the network can be separated into $M$ regions such that each agent can observe %
the nodes in its region. The objective is to control the voltage within a safety range while minimizing reactive power generation.
Accordingly, the shared reward at time step $t$ is
$ r^t = -\frac{1}{|V|}\sum_{i\in V} l_v(v_i) - \alpha \cdot l_q({\emph{q}^{PV})}       $
where $l_v(\cdot)$ represents the violation of voltage constraints and 
$l_q$ represents the reactive power generation loss. 

Variations of this model include an individual reward function for each agent that is locally computed based on the local measurements of the node and the neighboring nodes  \cite{gaoConsensusMultiAgentReinforcement2021}. Another variation uses a {\em Constrained MDP} (CMDP) that associates a cost to constraint violations and sets a budget that maintains nodal voltage profiles within a desirable range \cite{wang2019safe}.

\subsubsection{Solution Approaches}
 A prominent paradigm in \acrshort{marl} is that of \emph{Centralized Training Decentralized Execution (CTDE)} \cite{loweMultiAgentActorCriticMixed2017} by which agents expedite training by sharing their insights gained during training, i.e., sharing their leaned value functions, but operating according to individual policies. This approach is particularly relevant to \acrshort{psoc} settings in which agents are locally deployed, but their experiences and learned policies are relevant to nodes across the network.  
Accordingly, many \acrshort{psoc} frameworks adopt CTDE \cite{huMultiAgentDeepReinforcement2022,chenPhysicsShieldedMultiAgentDeep2022,liuOnlineMultiAgentReinforcement2021,liuRobustRegionalCoordination2021,gaoConsensusMultiAgentReinforcement2021,caoDataDrivenMultiAgentDeep2021}.  
The key differences between the approaches lie in the way the physical constraints are accounted for, the way scale is addressed, and the information that is assumed to be available for decision-making.

For example, in \cite{huMultiAgentDeepReinforcement2022}, the authors address the dual problem of \acrshort{pf} and \acrshort{vvc} at two different time scales. At a slower rate, \acrshort{pf} is resolved using a mixed-integer nonlinear programming method, while a model-free \acrshort{marl} approach works at high rate and aims to minimize voltage deviations of inverters and charging stations in the network. %
In \cite{chenPhysicsShieldedMultiAgentDeep2022} a {\em physics-shielding mechanism} penalizes agents for selecting actions that do not conform to a set of constraints. As the set of constraints is global, the centralized training parameters are shared between agents by sharing the learned {\em critic} that maintains the action values and is used to regulate the training process. 
In \cite{liuRobustRegionalCoordination2021} the network is partitioned into distinct, interconnected subnetworks, where each subnetwork is given agency over all of its \acrfull{pv} reactive power values. The reward for each agent in the environment is defined locally as the weighted sum of power loss and voltage violation of all the buses in the subnetwork.

\vspace{-.15cm}

 \subsection{Electricity Markets}
As opposed to traditional regulated markets where the role of each participant was distinct and fixed, modern energy market structures need to account for the dynamic nature of the market in which agents can control their consumption, production, and storage profiles in response to market signals and can interact with other agents and affect the grid's dynamics \cite{bose2019some,pinson2023future,capper2022peer,CHARBONNIER2022119188}.

Most traditional formulations of energy market problems such as \acrshort{uc}, \acrshort{ed} and \acrshort{dr} were adequate for centralized control of generation units and for unidirectional energy flow from transition to distribution. Solution methods for these problems typically rely on optimization methods such as mixed-integer linear programming (MILP) that are solved together with an interactive examination of the feasibility of the result using \acrshort{psoc} methods \cite{saravanan2013solution,zheng2014stochastic}.
With the emergence of novel smart grid structures and technologies that allow bidirectional energy flow and with the integration of active devices at the distribution leves, such methods are no longer sufficient.

Optimizing modern deregulated energy markets requires considering two perspectives. On the one hand, {\em mechanism design} is needed for the formation of economic mechanisms and protocols by which network agents can interact \cite{rosenschein1994rules,wooldridge2009introduction,shoham2008multiagent,leyton2022essentials}. %
This involves mechanisms such as auctions and monetary incentives to induce individual self-interested behaviors that maximize some global objective. On the other hand, there is a need to optimize the policies of the network agents given the underlying structure and mechanisms of interactions.

Importantly, unlike traditional marketplaces in which a key challenge is the predication of the behavior of strategic agents, here there is the additional need to account for the high number of heterogeneous participants and the various constraints imposed by the physics of the underlying electrical network and their impact on the market. %
Moreover, with the recent shift towards decentralized energy systems and the new information technologies, small-scale communities can coordinate their operation and satisfy more of their energy needs from local resources \cite{cremers2022efficient}. %
This raises the need to support various possibilities of local energy trading. %
However, despite various attempts at supporting these interactions, many governmental initiatives have failed due to the inability to support these highly dynamic and complex settings %
\cite{SCHWIDTAL2023113273,CAPPER2022112403}. 
Our focus is on harnessing \acrshort{marl} to achieve computational savings via distributed computations and to account for complex systems with strategic agents. 

\remove{
\cite{gaoConsensusMultiAgentReinforcement2021}
Power networks are a natural test field for MARL
algorithms. There are many problems in power networks involving the collaborative or competitive
actions of a vast number of agents, such as market bidding [5], voltage control, frequency control,
and emergency handling [6]. One of the obstacles for applying MARL in power networks is the lack
of transparency and guarantee, which is unacceptable considering the importance of reliable power
supply for the whole society. Therefore, it is sensible to start with problems that are less sensitive to
reliability but hard to be solved by conventional methods.
\cite{wooldridge2009introduction,shoham2008multiagent,leyton2022essentials}
}

\vspace{-.15cm}

\subsubsection{\acrshort{em} as a \acrshort{sg}}
Energy markets have many different forms and consider different optimization challenges: some consider network constraints while others rely on a post-process analysis to examine whether violations occur, some account for the cooperative and competitive nature of grid agents and their ability to adapt to the market dynamics while others use a fixed behavior (e.g., fixed consumption profile). This variety yielded many ways to model \acrshort{em}s as \acrshort{sg}s \cite{ZHU2023120212}.

Within this variety, many market settings, including \acrshort{uc}, \acrshort{dr} and \acrshort{ed}, can be modeled using two types of agents: a \acrfull{gm} that is responsible for managing the network and its markets and typically aims to maintain the network's stability while minimizing operation costs, and \acrfull{ga}s that represent market participants that aim to optimize their rewards while considering user-specific constraints and preferences (e.g., comfort, degradation costs, etc.). %

Assumption that control of generation is centralized, these settings can be solved in principle using a single-agent \acrshort{mdp}. However, even in such settings, it is often helpful to consider a decentralized formulation such as a Dec-POMDP \cite{bernstein2002complexity,oliehoek2016concise} to achieve distributed computation by representing each generation unit as a reward-maximizing agent. The challenge becomes finding a formulation that minimizes the information shared by agents to achieve optimal performance. In decentralized settings with self-interested agents modeled as a \acrshort{sg} there is the extra need to guarantee that the interaction mechanisms are {\em incentive-compatible} in that they induce truthful
reports by agents and that they achieve stability and social welfare \cite{shoham2008multiagent,leyton2022essentials}. This trade-off between local and global objectives is at the heart of \acrshort{marl} and multi-agent AI research.
\todo{decide if to elaborate here on the formulation In a fully decentralized version}

\remove{
For example, a \acrshort{sg} for the \acrshort{dr} problem considers \acrshort{gm} that can set electricity prices and the \acrfull{ga}s grid-edge agents (e.g., households) that can control their local storage units and decide when to sell their excess power to the grid. In contrast, in \acrshort{uc} and \acrshort{ed} settings, the focus is on managing and scheduling production of various generation units in response to consumption demands and the action space includes the dispatch of production commands at different time horizons. 
}

\remove{
In principle, under the assumption that control of the generation unit is centralized, these settings can be solved using a single-agent \acrshort{mdp}. %
In some settings, a centralized formulation is not relevant since \todo{complete}. However, even in centralized control settings, it is often helpful to consider a decentralized formulation to achieve distributed computation and reduce computational complexity. In a \acrshort{marl} version of this setting, each generation unit is represented as an agent with an individual reward function %
and the challenge becomes finding a suitable formulation that supports the global objective of minimizing the overall costs while respecting the operational constraints while minimizing the information that needs to be shared by the agents \cite{}. 

In decentralized settings in which agents may be self-interested, there is the extra need to make sure that the interaction mechanisms are {\em incentive-compatible} in that in that they induce truthful
reports by agents, and that they achieve stability and social welfare \cite{wooldridge2009introduction,shoham2008multiagent,leyton2022essentials}. This trade-off between local and global objectives is at the heart of \acrshort{marl} and multi-agent AI research.
\todo{decide if to elaborate here on the formulation In a fully decentralized version}.  
}

\vspace{-.15cm}

\subsubsection{Solution Approaches}
In general, it is challenging to design \acrshort{rl} and \acrshort{marl} solutions for problems that involve non-stationary and partially observable environments with self-interested agents \cite{huDistributedMultiagentReinforcement2023}. 
This is especially true for energy networks where demand and supply are hard to predict, there are many coupling constraints to consider, and the system is distributed. Thus, while this research area is on the rise \cite{ZHU2023120212}, there is currently limited work on using \acrshort{marl} for solving many of the \acrshort{em} problems and there are many open challenges for the integration of \acrshort{marl} for these problems. 

One recent example is \cite{charbonnierScalableMultiagentReinforcement2022} which presents a distributed \acrshort{rl} approach where agents collaborate (and share information) to regulate their consumption and optimize costs. 
Another example is the use of {\em distributed training with distributed execution} (DTDE) approach in which agents jointly estimate the total power demand using a  distributed communication protocol, locally decide their power generation value, and use another distributed computation to assess the total cost, which serves as the reward signal in the learning process 
\cite{huDistributedMultiagentReinforcement2023}. While this trend is on the rise \cite{ZHU2023120212}, there are many open challenges for the integration of \acrshort{marl} for these problems.

\remove{Another popular recent paradigm are multi-layered markets in which multiple markets at the bottom level
are cleared internally. An aggregator within each of these markets then
participates in a higher level market to clear excess supply or demand in the lower level markets. Multi-layer markets are often resolved with both single and double auctions.

An exception is the {\em distributed training with distributed execution} (DTDE) approach offered in \cite{huDistributedMultiagentReinforcement2023} for both continuous power demands and power outputs \sk{the problem being solver is referred to as the \acrfull{dedp} - what is the difference ?}. In this work, agents jointly estimate the total power demand using a  
distributed communication protocol, locally decide their power generation value, and use another distributed computation to assess the total cost, which serves as the reward signal in the learning process that is based on an actor-critic framework. While evaluation is reported as successful in terms of the ability to learn optimal joint generation policies and robustness to failure, the examined settings are limited in scale (up to ten generation units) and over a short horizon (up to seven time slots). 
}

\remove{First, the global power demand and cost are hard to predict locally. Second, it is difficult to consider the coupling constraints that need to be respected to obtain a feasible power
output distribution. 
In the absence of a centralized controller, agents are required to coordinate their actions in a
distributed way to satisfy the constraints. Third, it is
difficult to share the value function and policy 
in a distributed setting.}

\tm{A relevant MARL paper for DR we've done is https://doi.org/10.1016/j.apenergy.2022.118825}
\remove{
In several publications, namely \cite{shojaeighadikolaeiDemandResponsiveDynamic2020}, \cite{ghasemiMultiAgentDeepReinforcement2020}, and \cite{shojaeighadikolaeiDistributedEnergyManagement2022}, a joint \acrshort{dr}-\acrshort{dem} problem is introduced, wherein the grid manager is subjected to an additional quadratic cost for the utilization of reserve generations (only in \cite{shojaeighadikolaeiDistributedEnergyManagement2022}). The aforementioned problem is addressed through a decentralized, no-communication, \acrshort{dqn}-based \acrshort{marl} algorithm. Another publication, namely \cite{nweyeMERLINMultiagentOffline2022}, employs a decentralized \acrshort{sac} controller to solve the CityLearn \acrshort{marl} environment that deals with the \acrshort{dr} problem from a residential perspective. The authors explore various reward functions that incorporate local and global terms, such as electricity cost, carbon emission, power ramping, and others, which are supported by the CityLearn package. 

In \cite{chungDistributedDeepReinforcement2021}, the authors compare two different \acrshort{marl} control approaches for \acrshort{dr}, namely a centralized \acrshort{ddpg} agent called C-\acrshort{ddpg}, and D-\acrshort{ddpg}, a semi-decentralized \acrshort{ddpg} controller with a centralized critic and a policy network for each agent. Both methods are shown to converge to a \acrfull{ne}. Furthermore, the paper introduces consumers who have control over their shiftable-load scheduling (as opposed to prosumers), who are rewarded with a load satisfaction factor provided within the scheduling horizon. In \cite{zhangLearningBasedPowerManagement2020}, a \acrshort{dr} problem in the form of a \acrfull{mv} - \acrfull{lv} network is presented and modeled as a \acrshort{pomg}. A centralized Q-learning controller (aggregator) is employed for the \acrshort{mv} level, while the microgrids at all \acrshort{lv} levels maintain \acrfull{pf} simulations. Thus, by manipulating the electricity prices, the aggregator influences the \acrshort{pf} simulations results of its microgrid descendants and is rewarded based on their total revenues.
}

\remove{
After a mechanism is established, agents need to optimise their behavior in the given environment. 
In \cite{yangRecurrentDeepMultiagent2018}, the problem of setting optimal electricity tariffs in an electricity tariff market is addressed, with brokers that facilitate the buying and selling of electrical energy among consumers and producers by employing various real-time pricing strategies. Each broker is modeled as a \acrshort{mdp}, resulting in the electricity tariff market being modeled as a \acrshort{mamdp} implicitly. In this framework, each broker has control over the tariff for the next time step, aiming to maximize their revenue. To achieve this, the \acrfull{rdmrl} algorithm is used, which employs a reward shaping mechanism at its core. This mechanism localizes the global reward function for each broker by subtracting the irrelevant part for the $i^{th}$ broker from the global reward function, resulting in a set of local rewards that can be employed in a local manner.
}

\vspace{-.22cm}
\section{Research Gaps and Open Challenges}\label{sec:gaps}
\remove{The generality and flexibility \acrshort{marl} %
and its ability to support various multi-agent interactions and market mechanisms has led to various research threads that use \acrshort{marl} to support the decentralization of energy networks \cite{7779156,pub:33009,roche2010multi,ZHU2023120212,nair2018multi,mahela2020comprehensive,CAPPER2022112403}.} \remove{However, within the growing body of literature, most work so far was performed by non-AI researchers and lacks a principled formal account of the proposed frameworks.} 

To promote the use of MARL for \EnergyNetwork s we highlight the following research gaps that should be addressed. %

{\bf Gap 1 -- Inconsistent and non-unified \EnergyNetwork~problem definitions:}
While various %
frameworks have been suggested, they typically address a specific aspect of the system. This, together with the lack of standard definitions for \EnergyNetwork s and solution evaluation criteria, prevents the unified and standardized evaluation of proposed MARL approaches and hinders research progress.
    
{\bf Gap 2 -- Limited robustness, scalability and generalisability of MARL solutions:} %
While a range of MARL-based strategies for \EnergyNetwork~management have been developed,  it is unclear whether these will scale and perform well beyond the specific case studies they were trained for.
One current limitation is the reliance on strong simplifying assumptions, such as that  the environment is stationarity, whereas \EnergyNetwork s and markets are non-stationary and unpredictable. %
Also, evaluations of the suggested frameworks are typically done on small-scale settings which are far from realistic. 
Overcoming these challenges is key to the adoption of such methods by power system operators, especially given the stringent power sector reliability requirements.

{\bf Gap 3 -- Limited real-world data:} MARL methods typically require vast amounts of data and many environment interactions to train efficiently and to avoid any bias during training.
However, a limited volume of data can be  currently collected from real-world \EnergyNetwork s and used for training. 
This limits the generalisability and ability of the applied approaches to produce efficient policies, which is the specific advantage we seek in applying MARL to energy systems. 
This urges the need to find robust and scalable MARL solutions that work with limited data and to find ways to produce large volumes of high-quality data.

{\bf Gap 4 --  Lack of standardized simulations.} %
One way to address settings with limited data is by creating high-fidelity simulators that allow generating high-quality data. This is challenging in energy networks due to the need to account for both the strategic behavior of agents, as well as for the dynamics of the physical system. This is exacerbated by the integration of \RES s and \DER s that are highly uncertain and hard to predict and accurately model.  
Thus, while several recent MARL environments have been created to model certain aspects of grid management (e.g., \cite{pigott2022gridlearn,10.1145/3360322.3360998}), they do not offer a unified perspective on the physical considerations of the network together with the management of its induced market dynamics. %

\vspace{-.22cm}
\section{Conclusion}\label{sec:conclusion}
Recent advancements in \acrshort{marl} research and the availability of cost-effective computing have promoted the application of 
\acrshort{marl} methods to the effective management of energy systems. 
While this has resulted in many publications over the last decade, we believe many of the potential contributions are yet to be explored. A key insight from our exploration is that most work so far has been performed by power system researchers who use existing and possibly non-optimal \acrshort{marl} frameworks to solve complex energy network problems. On the flip side, a key barrier for AI researchers to become involved in these problems is that they typically lack the necessary domain knowledge and expertise to fully understand the critical challenges of energy network management. We hope that our review here will facilitate future collaboration efforts between researchers from both fields and will help unlock the full potential of using \acrshort{marl} to promote more efficient and sustainable energy networks.

\bibliographystyle{named}
\bibliography{SG_library}

\end{document}